\documentclass[conference,a4paper]{APSIPA2012}
\usepackage{multirow}
\usepackage[dvips]{graphicx}
\usepackage{amsmath}
\usepackage[psamsfonts]{amssymb}
\usepackage{amsxtra}
\usepackage{threeparttable}
\usepackage{graphicx,color}
\usepackage{amssymb,amsmath,bm}
\usepackage{textcomp}
\usepackage{array}
\usepackage{stfloats}
\usepackage{epsfig,epstopdf}

\begin{document}

\title{System Combination for Short Utterance Speaker Recognition}

\author{%
\authorblockN{%
Lantian Li\authorrefmark{2}, Dong Wang\authorrefmark{2}, Xiaodong Zhang\authorrefmark{3}, Thomas Fang Zheng$^*$\authorrefmark{2}, Panshi Jin\authorrefmark{3}
}

\authorblockA{%
\authorrefmark{2}
Center for Speech and Language Technologies, Division of Technical Innovation and Development, \\
Tsinghua National Laboratory for Information Science and Technology;\\
Center for Speech and Language Technologies, Research Institute of Information Technology;\\
Department of Computer Science and Technology, Tsinghua University, Beijing, China\\
}

\authorblockA{%
\authorrefmark{3}
IT Management Department, China Construction Bank}

\authorblockA{%
\authorrefmark{1}
Corresponding Author E-mail: fzheng@tsinghua.edu.cn} \\

}
\maketitle

\thispagestyle{empty}

\begin{abstract}

For text-independent short-utterance speaker recognition (SUSR), the performance often degrades dramatically.
This paper presents a combination approach to the SUSR tasks with two phonetic-aware systems:
one is the DNN-based i-vector system and the other is our recently proposed subregion-based GMM-UBM system.
The former employs phone posteriors to construct an i-vector model in which the shared statistics offers stronger robustness against limited test data, while the latter establishes a phone-dependent GMM-UBM system which represents speaker characteristics with more details.
A score-level fusion is implemented to integrate the respective advantages from the two systems.
Experimental results show that for the text-independent SUSR task, both the DNN-based i-vector system and the subregion-based GMM-UBM system outperform their respective baselines,
and the score-level system combination delivers performance improvement.
\end{abstract}

\section{Introduction}

After decades of research, current text-independent speaker recognition (SRE) systems can obtain rather
good performance, if the test utterances are sufficiently long~\cite{bimbot2004tutorial,Kinnunen10,greenberg20132012}.
However, if the utterances are short, serious performance degradation is often observed.
For instance, Vogt et al.~\cite{vogt2010making} reported that when the test speech was shortened
from $20$ seconds to $2$ seconds,
the performance degraded sharply in terms of equal error rate (EER) from 6.34\% to 23.89\% on a NIST SRE task.
The performance degradation seriously limits
the application of SRE in practice, since long-duration test would impact user experience
significantly, and in many situations it is very difficult, if not possible, to collect sufficient
test data, for example in forensic applications.
How to improve performance of speaker recognition on short utterances (SUSR) is an open research topic.

A multitude of studies have been conducted in SUSR. For example, in~\cite{vogt2008factor}, the authors
showed that the performance on short utterances can be improved by JFA. This work was extended in~\cite{kanagasundaram2011vector}
which reported that the i-vector model can distill speaker information in a more effective way so it
is more suitable for SUSR.
In addition, a score-based segment selection technique was proposed in~\cite{nosratighods2010segment}.
A relative EER reduction of 22\% was
reported by the authors on a recognition task where the test utterances were shorter than $15$
seconds in length.

We argue that the difficulty associated with text-independent SUSR can be largely attributed to the
mismatched distributions of speech data between enrollment and test.
Assuming that the enrollment speech is sufficiently long, so the speaker model can be well trained.
If the test speech is sufficient as well, the distribution of the test data tends to match the
distribution represented by the speaker model; however, if the test speech is short, then only a part
of the probability mass represented by the speaker model can be covered by the test speech.
For a GMM-UBM system, this is equal to say that only a few Gaussian components of the model
are covered by the test data, and therefore the likelihood evaluation is biased.
For the i-vector model, since the Gaussian components share some statistics via a single
latent variable, the impact of short test speech is partly alleviated. However, the limited
data anyway leads to insufficient evaluation of the Baum-Welch statistics, resulting in
a less reliable i-vector inference.

A possible solution for the text-independent SUSR problem is to identify the phone content
of the speech signal, and then model and evaluate speakers on individual phones. We
call this `phonetic-aware' approach. This
approach can be regarded as a transfer from a text-independent task to a text-dependent task.
The latter is certainly more resilient to short utterances, as has been demonstrated in~\cite{larcher2012rsr2015}.

Two phonetic-aware approaches have been proposed.
One is the subregion model based on the GMM-UBM architecture~\cite{li2016susr}, and the other is the DNN-based i-vector model~\cite{lei2014novel,kennydeep}.
Both the two approaches employ an automatic speech recognition
(ASR) system to generate phone transcriptions or posteriors for enrollment speech, and then establish a
phonetic-aware speaker model based on the transcriptions or posteriors.
These two approaches, however, are different
in model structure and implementation.
The subregion modeling approach builds multiple phone-dependent UBMs and speaker GMMs,
and evaluates test speech on the phone-dependent models.
The DNN-based i-vector approach, in contrast, keeps the single UBM/GMM framework,
but relates each Gaussian component to a phone or a phone state.
The former tends to be more flexible when learning speaker characteristics, while the latter
is more robust against limited test data, due to the low-dimensional latent
variable that is shared among all the Gaussian components.
We therefore argue that the two approaches can be combined, so that
the respective advantages of the two methods can be integrated.

The rest of the paper is organized as follows: Section~\ref{sec:rel} discusses some
related work, Section~\ref{sec:sbm} presents
the subregion model, and Section~\ref{sec:frame} describes the combination approach.
Section~\ref{sec:exp} presents the experiments, and the entire paper is concluded in Section~\ref{sec:conl}.

\section{Related work}
\label{sec:rel}

The idea of employing phonetic information in speaker recognition has been investigated
by previous research studies. For instance, Omar et al.~\cite{omar2010training} proposed to derive UBMs from
Gaussian components of a GMM-based ASR system, with a K-means clustering approach based on the symmetric KL
distance. The DNN-based i-vector method was proposed in ~\cite{lei2014novel,kennydeep}.
In the work, posteriors of senones (context-dependent states) generated by a DNN trained for ASR were used for
model training as well as i-vector inference.
Note that all these studies focus on relatively long utterances (5-10 seconds), whereas  our study
in this paper focuses on utterances as short as $0.5$ seconds.

\section{Subregion modeling}
\label{sec:sbm}

We briefly describe the subregion model presented by us recently~\cite{li2016susr}.
The basic idea is firstly presented, and then the implementation details are described.

\subsection{Acoustic subregions}

The conventional GMM-UBM system treats the entire acoustic space as a whole
probabilistic space, and computes the likelihood of an input speech signal
by a GMM model, formulated as follows:

\vspace{-2mm}
\[
p(x;s)   =   \prod_{t} \sum_{c} P(c) \mathcal{N}(x_t;\mu^s_{c}, \Sigma_{c})
\]
\vspace{-2mm}

\noindent where $x$ denotes the speech signal, and $\mathcal{N}(x; \mu, \Sigma)$
represents a Gaussian distribution with $\mu$ as the mean and $\Sigma$ as the covariance matrix.
Further more, $c$ indexes the Gaussian component, and $s$ indexes the speaker. $P(c)$
is a prior distribution on the $c$-th component. Roughly speaking,
this model splits the acoustic space into a number of subregions, and each subregion
is modelled by a Gaussian distribution.

There are at least three potential problems with this model: (1) the
subregion splitting is based on unsupervised clustering (via the EM algorithm~\cite{moon1996expectation}),
so it is not necessarily meaningful in phonetic; (2) each subregion is
modeled by a Gaussian, which seems too simple; (3) the priors over the
subregions are fixed, independent of $x_t$.

The subregion model was proposed to solve these problems. Firstly, the acoustic
space is split into subregions that roughly correspond to phonetic units (e.g., phones);
secondly, each subregion is modelled by a GMM instead of a single Gaussian; thirdly, the weight for
each subregion is based on the posterior $P(c|x_t)$ instead of the prior $P(c)$.
This is formulated as follows:

\vspace{-2mm}
\[
p(x;s)   =   \prod_{t} \sum_{c} P(c|x_t) \sum_{k} \pi_{c,k} \mathcal{N}(x_t;\mu^s_{c,k}, \Sigma_{c,k})
\]
\vspace{-2mm}

\noindent where $k$ indexes the Gaussians within a subregion GMM. A key component of this
model is the posterior probability $P(c|x_t)$, which is not a pre-trained constant value, but an
assignment of each signal $x_t$ to the subregions. In our study, this quantity is generated by
an ASR system.

\subsection{Speech units}

The inventory of speech units varies for different languages. In Chinese, the language focused in this
paper, Initials and Finals (IF) are the most commonly used~\cite{zhang2001improved}.
Roughly speaking, Initials correspond to consonants, and Finals correspond to vowels and nasals.
Among the IFs, Finals are recognized to convey more speaker
related information~\cite{beigi2011fundamentals, gongcheng2014}, and therefore are used
as the speech units in this study.

Using Finals to train the subregion model is not very practical, because
there is a large number of Finals, and most Finals can only find limited
data in both training and test.
A possible solution is to cluster similar units together and build subregion
models based on the resulting speech unit classes.
In this study, we develop a vector quantization (VQ) method based on the
K-means algorithm to conduct the clustering.

\subsection{Subregion modeling based on speech unit classes}

Denote the speech unit classes (Final clusters) by $\{$SUC-$c$,$c$$=1,...,C\}$,
a subregion UBM can be trained for each SUC-$c$ with the training data that are aligned to
the Finals in SUC-$c$ by the ASR system.
The subregion UBM of class SUC-$c$ is denoted by $\lambda^{UBM}_c$. The speaker-dependent
subregion GMMs can be trained based on the subregion UBMs,
using the enrollment data that have been aligned to the Finals of each cluster.







Once the speaker-dependent subregion GMMs are trained, a test utterance can be scored
on each subregion.
Suppose a test utterance contains $L$ Finals according to the decoding result of speech recognition,
and denote the speech unit class of the $l$-th Final by $c(l)$. Further denote the speech segment
of this unit by $X_l$, and its length is $T_l$.
The score of $X_l$ is measured by the log likelihood ratio between the subregion speaker-dependent
GMM $\lambda^s_{c(l)}$ and the subregion UBM $\lambda^{UBM}_{c(l)}$,
where $s$ denotes the speaker. This is formulated by:

\vspace{-2mm}
\[
   \varphi_{s, l} = \log p (\mathbf{X}_{l}|\lambda_{c(l)}^{s}) - \log p (\mathbf{X}_{l}|\lambda_{c(l)}^{ubm} )
\]
\vspace{-2mm}

\noindent Finally, the score of the entire utterance can be computed as the average of the subregion-based scores.


\section{System combination}
\label{sec:frame}

In this section, we first describe the difference between the subregion model and another
phonetic-aware method: the DNN based i-vector model. Then the combination system is presented.

\subsection{DNN-ivector and subregion model}

The DNN-based i-vector approach proposed by Lei and colleagues~\cite{lei2014novel} replaces
GMM-based posteriors by DNN-generated posteriors when computing the Baum-Welch statistics
for model training and i-vector inference.
The DNN model is trained for speech recognition, so the output targets correspond to phones or states.
This essentially builds a UBM and speaker GMMs where the Gaussian components
correspond to phones or states. This
is quite similar as the subregion model, though the
model structures of the two models are different.
On one hand, the subregion model builds GMMs for each
subregion, while the
DNN-based i-vector approach still assumes Gaussian for each subregion. From this aspect,
the subregion model tends to be more flexible and represents speaker characteristics with more details.
On the other hand, the subregions in the subregion modeling are relatively independent, whereas
the subregions in the DNN-ivector model share statistics via the latent variable (i-vector).
This sharing may lead to more strong robustness against limited test data.

\subsection{Score-level system combination}

\begin{figure}[htp]
\begin{center}
\includegraphics[width=0.9\linewidth]{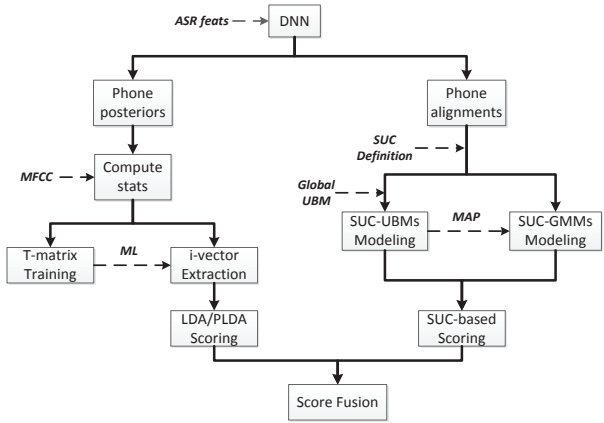}
\end{center}
\caption{The diagram of the score-level system combination.}
\label{fig:frame}
\end{figure}

Due to the difference of the two phonetic-aware models and their prospective advantages, it is reasonable to combine them together.
The combination system involves three components. Firstly, a DNN model for ASR
is trained and used to generate the phonetic information: phone posteriors and phone alignments.
Secondly, the phone posteriors are used to train the DNN-based i-vector model,
and the phone alignments are used to build the subregion model.
Thirdly, when scoring a test speech, the scores derived
from the DNN-ivector system and the
subregion GMM-UBM system are averaged
to make the final decision. Fig.~\ref{fig:frame} illustrates the system framework.

\section{Experiments}
\label{sec:exp}

\subsection{Database}

\subsubsection{Database for evaluation (SUD12)}

There is not a standard database for performance evaluation on text-independent SUSR tasks.
Therefore, we firstly designed and recorded a database that is suitable for SUSR research and published it for research
usage\footnote{http://www.cslt.org/resources.php?Public\%20data}.
The database was named as ``SUD12''~\cite{li2016susr}~\cite{zhang2012k}, and was designed in the principle to guarantee sufficient IF coverage.
In order to focus on short utterances and exclude other factors such as the channel and emotion, the recording was
conducted in the same room and with the same microphone, and the reading style was neutral. The database
consists of $28$ male speakers and $28$ female speakers, and all the utterances are in standard Chinese.
For each speaker, there are $100$ Chinese sentences, each of which contains $15$ $\sim$ $30$ Chinese characters.
The sampling rate is $16$ kHz with $16$-bits precision.

The enrollment database involves all the $56$ speakers. For each speaker, the effective speech signals for enrollment is about $35$ seconds.
The test database consists of $56$ speakers, and each speaker involves $62$-$63$ short utterances that cover
all the Finals.
The length of each utterance is not more than $2$ seconds and mostly as short as $0.5$ seconds.
With the test database, $3,523$ target trials and $197,293$ non-target trials are defined for performance evaluation.

\subsubsection{Database for UBM training (863DB)}

The speech data used to train the UBMs, subregion UBMs and T-matrix were chosen from the 863 Chinese speech corpus.
The 863 database was well designed to cover all the Chinese IFs, so it is particularly suitable to train subregion UBMs based on Final classes.
All the recordings are at a sampling rate of $16$ kHz, and the sample precision is $16$ bits. In this study,
we choose $17$ hours of speech data and denote the database by 863DB.

\subsection{Experimental conditions}

The Kaldi toolkit~\cite{Povey_ASRU2011} was used to conduct the experiments.
Following the standard recipe of SRE08, the acoustic feature was the
conventional $60$-dimensional Mel frequency cepstral coefficients (MFCCs),
which involved $20$-dimensional static components plus the first and
second order derivatives. The frame size was $25$ ms and the frame shift was $10$ ms.
Besides, a simple energy-based voice activity detection (VAD) was performed before the feature extraction.

The ASR system used to generate the phone alignment was a large-scale DNN-HMM hybrid
system. The system was trained using Kaldi following the WSJ S5 recipe.
The feature used was 40-dimensional Fbanks. The basic features were spliced by
a window of $11$ frames, and an LDA (linear discriminative analysis)
transform was applied to reduce the dimensionality to $200$.
The DNN structure involved $4$ hidden layers, each containing $1,200$ hidden
units. The output layer contained $6,761$ units, corresponding to the number of GMM senones.
The DNN was trained with $6,000$ hours of speech signals, and the decoding
employed a powerful $5$-gram language model trained on $2$ TB text data.

We chose the conventional GMM-UBM approach to construct the baseline SUSR system.
The UBM consisted of $1,024$ Gaussian components and was trained with the 863DB.
The SUD12 was employed to conduct the evaluation. With the enrollment data,
the speaker GMMs were derived from the UBM by MAP, where the MAP adaptation
factor was optimized so that the EER on the test set was the best.
For comparison, a GMM-based i-vector system was also constructed.
The training was based on the same UBM model as the GMM-UBM system,
and the dimensionality of the i-vector was $400$.

For the DNN-based i-vector system, the DNN model was trained following
the same procedure as the one used for the ASR system, but with less
number of senones. In our experiments, the number was $928$, comparable
to the number of Gaussian components of the GMM-UBM
system.
The dimensionality of the DNN-based i-vectors was set to $400$.

\subsection{Basic results}

We first investigated the subregion model based on speech unit classes.
For this model, the number of speech unit classes need to be defined before hand.
In our experiments, we observed that either too small or too large
clustering numbers lead to suboptimal performance, and the optimal
setting in our experiment was $C$=6~\cite{li2016susr}.
Table I shows the derived unit classes. It can be seen
that the resultant clusters are intuitively reasonable.

\vspace{-2mm}
\begin{table}[htb]
\normalsize
\begin{center}
\caption{Speech unit classes derived by k-means clustering.}
\begin{tabular}{l|c}
\hline
Class & Speech Units \\
\hline
\hline
1 & a, ao, an, ang, ia, iao, ua\\
2 & e, ei, ai, i, ie, uei, iii\\
3 & iou, ou, u, ong, uo, o\\
4 & v, vn, ve, van, er\\
5 & en, ian, uan, uen, uai, in, ii, ing\\
6 & eng, iang, iong, uang, ueng\\
\hline
\end{tabular}
\end{center}
\label{tab:cluster}
\vspace{-4mm}
\end{table}

The results in terms of EER are presented in Table II, where `GMM-UBM' is the GMM-UBM baseline system,
`SBM-DD' denotes the subregion modeling system ($C$=6).
`GMM i-vector' denotes the traditional GMM-based i-vector system with cosine distance metric,
and `DNN i-vector' denotes the DNN-based i-vector system with cosine distance metric.

We first observe that both the subregion modeling system and the DNN-based i-vector system
outperform their relative baselines (`GMM-UBM' and `GMM-based i-vector') in a significant way.
This confirms the effectiveness of the two phonetic-aware methods.
Besides, it can be seen that the GMM-UBM baseline outperforms the two i-vector systems,
but after the probabilistic linear discriminant analysis (PLDA)~\cite{prince2007probabilistic} is employed,
the i-vector system is improved and outperforms the GMM-UBM system.

\begin{table}[htb]
\normalsize
\begin{center}
\caption{Performance of phonetic-aware methods}
\begin{tabular}{l|c}
\hline
System     & EER (\%)      \\
\hline
\hline
GMM-UBM     & 28.97\\
SBM-DD     & 22.74 \\
\hline
\hline
GMM i-vector        & 39.91 \\
DNN i-vector          & 29.61 \\
DNN i-vector + PLDA         & 19.16 \\
\hline
\hline
Combination system   &   17.43  \\
\hline
\end{tabular}
\end{center}
\vspace{-4mm}
\label{tab:subregion}
\end{table}



\subsection{System combination}

\begin{figure}[!htb]
\begin{center}
\includegraphics[width=0.8\linewidth]{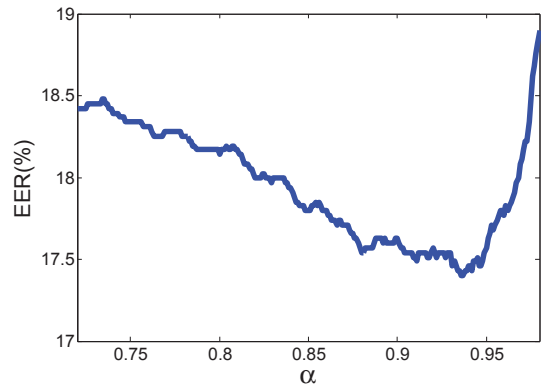}
\end{center}
\vspace{-2mm}
\caption{Performance of score-level system combination with the interpolation factor $\alpha$.}
\label{fig:fusion}
\end{figure}

We combine the `DNN i-vector + PLDA' system and the `SBM-DD' system by a linear score fusion:
$\alpha s_{plda} + (1-\alpha) s_{sbm}$, where $\alpha$ is the interpolation factor.
Fig.~\ref{fig:fusion} presents the performance with various $\alpha$. It clearly shows that
the system combination leads to better performance than each individual system.
Fig.~\ref{fig:fusion} shows that $\alpha$=$0.94$ is a good choice.
Table II has shown the results
of the combination system with this configuration.

\section{Conclusions}
\label{sec:conl}

This paper presents a combination system to deal with short utterances in text-independent speaker recognition.
This system combines two phonetic-aware methods: one is the DNN-based i-vector system
and the other is the subregion-based GMM-UBM system.
The experimental results show that both the DNN-based i-vector system and the subregion-based
GMM-UBM system outperforms their respective baselines,
and a simple score fusion leads to the best performance we have obtained so far.
The strategy presented in this paper has been verified in and applied to the Mobile banking of CCB (China Construction Bank) and has been achieving good performance.
Future work involves combination of feature-based and model-based
compensations for short utterances, and investigation on
phone-discriminative methods.

\section*{Acknowledgment}

This work was supported by the National Natural Science Foundation of China under Grant No. 61371136 and No. 61271389, it was also supported by the National Basic Research Program  (973 Program) of China under Grant No. 2013CB329302.

\bibliographystyle{IEEEtran}
\bibliography{susr}

\end{document}